\pdfoutput=1
\documentclass[runningheads]{llncs}
\usepackage{makeidx}  
\usepackage{gensymb}
\usepackage{bm}
\usepackage{graphicx}
\usepackage[hidelinks]{hyperref}

\begin{document}
\frontmatter          
\pagestyle{headings}  
\addtocmark{} 

\mainmatter

\title{Domain-adversarial neural networks \newline to address the appearance variability of histopathology images}

\titlerunning{DANNs to address the appearance variability of histopathology images}

\author{Maxime W. Lafarge \and Josien P.W. Pluim \and Koen A.J. Eppenhof \and Pim Moeskops \and Mitko Veta}
\authorrunning{M.W. Lafarge et al.} 

\institute{Medical Image Analysis Group, Department of Biomedical Engineering,\\
Eindhoven University of Technology, The Netherlands}

\maketitle 

\begin{abstract}
Preparing and scanning histopathology slides consists of several steps, each with a multitude of parameters. The parameters can vary between pathology labs and within the same lab over time, resulting in significant variability of the tissue appearance that hampers the generalization of automatic image analysis methods. Typically, this is addressed with ad-hoc approaches such as staining normalization that aim to reduce the appearance variability.
In this paper, we propose a systematic solution based on domain-adversarial neural networks. We hypothesize that removing the domain information from the model representation leads to better generalization. We tested our hypothesis for the problem of mitosis detection in breast cancer histopathology images and made a comparative analysis with two other approaches.
We show that combining color augmentation with domain-adversarial training is a better alternative than standard approaches to improve the generalization of deep learning methods.
\end{abstract}

\section{Introduction}
Histopathology image analysis aims at automating tasks that are difficult, expensive and time-consuming for pathologists to perform.
The high variability of the appearance of histopathological images, which is the result of the inconsistency of the tissue preparation process, is a well-known observation.
This hampers the generalization of image analysis methods, particularly to datasets from external pathology labs.

The appearance variability of histopathology images is commonly addressed by standardizing the images before analysis,
for example by performing staining normalization \cite{ruifrok2001quantification,macenko2009method}.
These methods are efficient at standardizing colors while keeping structures intact, but are not equipped to handle other sources of variability, for instance due to differences in tissue fixation. 

We hypothesize that a more general and efficient approach in the context of deep convolutional neural networks (CNNs) is to impose constraints that disregard non-relevant appearance variability with domain-adversarial training \cite{ganin2016domain}.
We trained CNN models for mitosis detection in breast cancer histopathology images on a limited amount of data from one pathology lab and evaluated them on a test dataset from different, external pathology labs.
In addition to domain-adversarial training, we investigated two additional approaches (color augmentation and staining normalization) and made a comparative analysis.
As a main contribution, we show that domain-adversarial neural networks are a new alternative for improving the generalization of deep learning methods for histopathology image analysis.

\section{Materials and Methods}
\subsection{Datasets}
This study was performed with the TUPAC16 dataset \cite{tupac2016} that includes 73 breast cancer cases annotated for mitotic figures.
The density of mitotic figures can be directly related to the tumor proliferation activity, and is an important biomarker for breast cancer prognostication. 

The cases come from three different pathology labs (23, 25 and 25 cases per lab)
and were scanned with two different whole-slide image scanners (the images from the second two pathology labs were scanned with the same scanner).
All CNN models were trained with eight cases (458 mitoses) from the first pathology lab.
Four cases were used as a validation set (92 mitoses).
The remaining 12 cases (533 mitoses) from the first pathology lab were used as an internal test set\footnote{This test set is identical to the one used in the AMIDA13 challenge \cite{veta2015assessment}.}
and the 50 cases from the two other pathology labs (469 mitoses) were used to evaluate inter-lab generalization performance.

\subsection{The Underlying CNN Architecture}
\label{ref:baselineCNN}
The most successful methods for mitosis detection in breast cancer histopathology images are based on convolutional neural networks (CNN).
These methods train models to classify image patches based on mitosis annotations resulting from the agreement of several expert pathologists \cite{cirecsan2013mitosis,veta2015assessment,tupac2016}.

The baseline architecture that is used in all experiments of this study is a 6-layer neural network with four convolutional and two fully connected layers that takes a $63\times63$ image patch as an input and produces a probability that there is a mitotic figure in the center of the patch as an output.
The first convolutional layer has $4\times4$ kernels and the remaining three convolutional layers have $3\times3$ kernels.
All convolutional layers have 16 feature maps.
The first fully connected layer has 64 neurons and the second layer serves as the output layer with softmax activation.
Batch normalization, max-pooling and ReLU nonlinearities are used throughout.
This architecture is similar to the one proposed in \cite{cirecsan2013mitosis}.
The neural network can be densely applied to images in order to produce a mitosis probability map for detection.

\subsection{Three Approaches to Handling Appearance Variability}
\label{ref:approaches}
Poor generalization occurs when there is a discrepancy between the distribution of the training and testing data.
Increasing the amount of training data can be of help, however, annotation of histology images is a time-consuming process that requires scarce expertise.
More feasible solutions are needed, therefore we chose to investigate three approaches.

One straightforward alternative is to artificially produce new training samples.
Standard data augmentation methods include random spatial and intensity/color transformation (e.g. rotation, mirroring and scaling, and color shifts).
In this study, we use spatial data augmentation (arbitrary rotation, mirroring and $\pm20\%$ scaling) during the training of all models.
Since the most prominent source of variability in histopathology images is the staining color appearance, the contribution of color augmentation (CA) during training is evaluated separately.

The opposite strategy is to reduce the appearance variability of all the images as a pre-processing step before training and evaluating a CNN model.
For hematoxylin and eosin (H\&E) stained slides, staining normalization (SN) methods can be used \cite{ruifrok2001quantification,macenko2009method}.

A more direct strategy is to constrain the weights of the model to encourage learning of mitosis-related features that are consistent for any input image appearance.
We observed that the features extracted by a baseline CNN mitosis classifier carry information about the origin of the input patch (see Section~\ref{results}).
We expect that better generalization can be achieved by eliminating this information from the learned representation with domain-adversarial training \cite{ganin2016domain}.

Finally, in addition to the three individual approaches, we also investigate all possible combinations. 

\subsubsection{Color Augmentation.}
Color variability can be increased by applying random color transformations to original training samples.
We perform color augmentation by transforming every color channels $I_{c} \leftarrow a_{c} \cdot I_{c} + b_{c}$, where $a_c$ and $b_c$ are drawn from uniform distributions $a_{c} \sim U\left[0.9, 1.1\right]$ and $b_{c} \sim  U\left[-10,+10\right]$.

\begin{figure}[ht]
\label{fig:augmentations}
\centering
\includegraphics[width=1.0\textwidth, trim=25pt 325pt 60pt 25pt, clip]{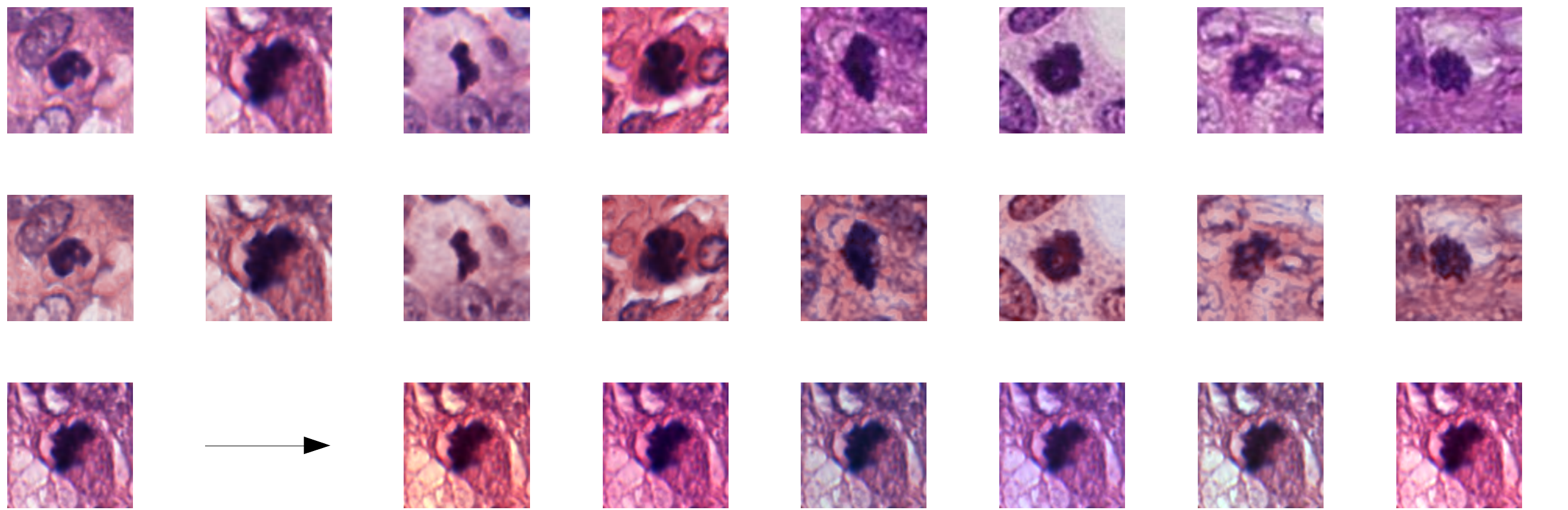}
\caption{Illustration of the variability of histological images with 8 patches from different slides (first row),
and their transformed version after staining normalization (second row).
The third row illustrates the range of color variation induced by color augmentation.}
\end{figure} 

\subsubsection{Staining Normalization.}
The RBG pixel intensities of H\&E-stained histo\-pathology images can be modeled with the Beer-Lambert law of light absorption: $I_{c} = I_{0} \exp\left(- \mathbf{A}_{c,*} \cdot \mathbf{C} \right)$. In this expression $c=1, 2, 3$ is the color-channel index, $\mathbf{A} \in [0,+\infty]^{3 \times 2} $ is the matrix of absorbance coefficients and $\mathbf{C} \in [0,+\infty]^{2}$ are the stain concentrations \cite{ruifrok2001quantification}.
We perform staining normalization with the method described in \cite{macenko2009method}.
This is an unsupervised method that decomposes any image with estimates of its underlying $\mathbf{A}$ and $\mathbf{C}$.
The appearance variability over the dataset can then be reduced by recomposing all the images using some fixed absorbance coefficients.

\subsubsection{Domain Adversarial Neural-Network.}
\label{ref:daNN}
Since every digital slide results from a unique combination of preparation parameters,
we assume that all the image patches extracted from the same slide come from the same unique data distribution and thus constitute a domain.
Domain-adversarial neural networks (DANN) allow to learn a classification task,
while ensuring that the domain of origin of any sample of the training data cannot be recovered from the learned feature representation \cite{ganin2016domain}.
Such a domain-agnostic representation improves the cross-domain generalization of the trained models.

Any image patch $\mathbf{x}$ extracted from the training data can be given two labels:
its class label $y$ (assigned to ``$1$'' if the patch is centered at a mitotic figure, ``$0$'' otherwise)
and its domain label $d$ (a unique identifier of the slide that is the origin of the patch).

\begin{figure}
\centering
\includegraphics[width=1.0\textwidth, trim=40pt 430pt 285pt 25pt, clip]{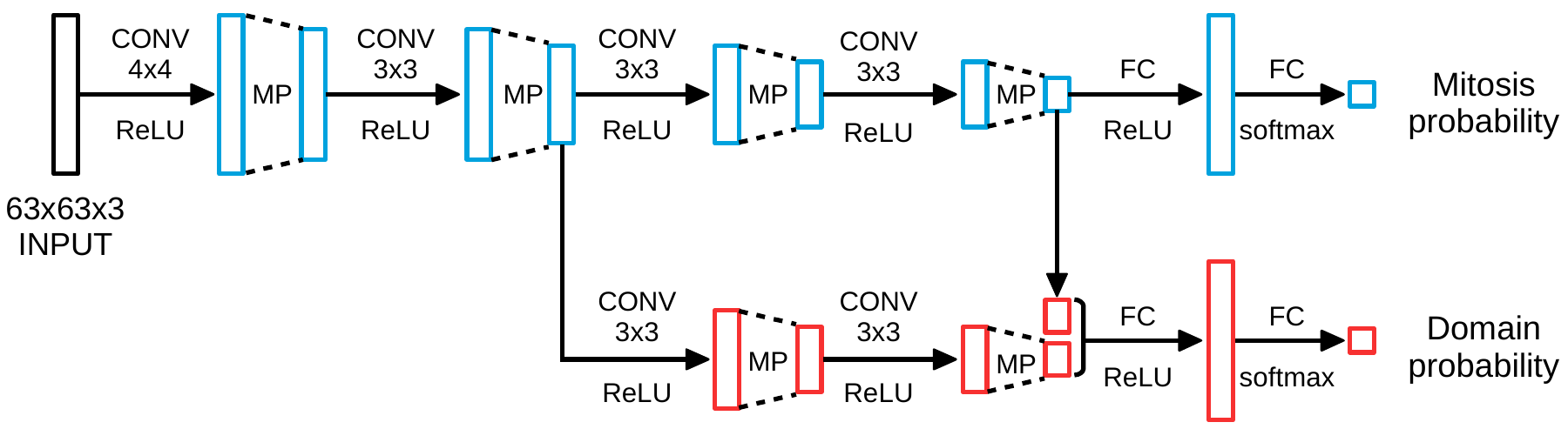}
\caption{
Architecture of the domain-adversarial neural network.
The domain classification (red) bifurcates from the baseline network (blue) at the second and forth layers.}
\label{fig:daNN}
\end{figure}

The training of the mitosis classifier introduced in Sect. \ref{ref:baselineCNN} is performed by minimizing
the cross-entropy loss $\mathcal{L}_{M}(\mathbf{x}, y; \boldsymbol{\theta}_{M})$, where $\boldsymbol{\theta}_{M}$ are the parameters of the network. 

The DANN is made of a second CNN that takes as input the activations of the second and fourth layers of the mitosis classifier and predicts the domain identifier $d$.
This network is constructed in parallel to the mitosis classifier, with the same corresponding architecture (Fig. \ref{fig:daNN}).
Multiple bifurcations are used to make domain classification possible from different levels of abstraction and to improve training stability as in \cite{kamnitsas2016domain}.
The cross-entropy loss of the domain classifier is $\mathcal{L}_{D}(\mathbf{x}, d; \boldsymbol{\theta}_{M}, \boldsymbol{\theta}_{D})$, where $\boldsymbol{\theta}_{D}$ are the parameters of the bifurcated network (note however that the loss is also a function of $\boldsymbol{\theta}_{M}$).

The weights of the whole network are optimized via gradient back-propagation during an iterative training process that consists of three successive update rules:
\begin{eqnarray} \label{eq:updateMitosis}
\begin{minipage}[l]{0.6\textwidth}
Optimization of the mitosis classifier \\ with learning rate $\lambda_{M}$:
\end{minipage}
&
\begin{minipage}[l]{0.35\textwidth}
$\boldsymbol{\theta}_{M} \leftarrow\boldsymbol{\theta}_{M} - \lambda_{M} \frac{\partial \mathcal{L}_{M}}{\partial \boldsymbol{\theta}_{M}}$
\end{minipage}
\end{eqnarray}
\begin{eqnarray} \label{eq:updateDomains}
\begin{minipage}[l]{0.6\textwidth}
Optimization of the domain classifier \\ with learning rate $\lambda_{D}$:
\end{minipage}
&
\begin{minipage}[l]{0.35\textwidth}
$\boldsymbol{\theta}_{D} \leftarrow \boldsymbol{\theta}_{D} - \lambda_{D} \frac{\partial \mathcal{L}_{D}}{\partial \boldsymbol{\theta}_{D}}$
\end{minipage}
\end{eqnarray}
\begin{eqnarray} \label{eq:updateAdversarial}
\begin{minipage}[l]{0.6\textwidth}
Adversarial update of the mitosis classifier:
\end{minipage}
&
\begin{minipage}[l]{0.35\textwidth}
$\boldsymbol{\theta}_{M} \leftarrow \boldsymbol{\theta}_{M} + \alpha \lambda_{D} \frac{\partial \mathcal{L}_{D}}{\partial \boldsymbol{\theta}_{M}}$
\end{minipage}
\end{eqnarray}

The update rules (\ref{eq:updateMitosis}) and (\ref{eq:updateAdversarial}) work in an adversarial way: 
with (\ref{eq:updateMitosis}), the parameters $\boldsymbol{\theta}_{M}$ are updated for the mitosis detection task (by minimizing $\mathcal{L}_{M}$),
and with (\ref{eq:updateAdversarial}), the same parameters are updated to prevent the domain of origin to be recovered from the learned representation (by maximizing $\mathcal{L}_{D}$).
The parameter $\alpha \in \left[0,1\right]$ controls the strength of the adversarial component.

\newcommand{\fscore}[1]{F\textsubscript{#1}-score}
\subsection{Evaluation}
The performances of the mitosis detection models were evaluated with the \fscore{1} as described in \cite{cirecsan2013mitosis,veta2015assessment,tupac2016}.
We used the trained classifiers to produce dense mitosis probability maps for all test images.
All local maxima above an operating point were considered detected mitotic figures.
The operating point was determined as the threshold that maximizes the \fscore{1} over the validation set.

We used the t-distributed stochastic neighbor embedding (t-SNE) \cite{maaten2008visualizing} method for low-dimensional feature embedding,
to qualitatively compare the domain overlap of the learned feature representation for the different methods.

\section{Experiments and Results}
\label{results}
For every possible combination of the three approaches developed in Section \ref{ref:approaches},
we trained three convolutional neural networks with the same baseline architecture,
under the same training procedure, but with random initialization seeds to assess the consistency of the approaches.

\subsubsection{Baseline Training.}
Training was performed with stochastic gradient descent with momentum and with the following parameters:
batch size of 64 (with balanced class distribution),
learning rate $\lambda_{M}$ of 0.01 with a decay factor of 0.9 every 5000 iterations,
weight decay of 0.0005 and momentum of 0.9.
The training was stopped after 40000 iterations.

Because the training set has a high class imbalance, hard negative mining was performed as previously described \cite{cirecsan2013mitosis,veta2016mitosis}.
To this purpose, an initial classifier was trained with the baseline CNN model.
A set of hard negative patches was then obtained by probabilistically sampling the probability maps produced by this first classifier (excluding ground truth locations).
We use the same set of hard-negative samples for all experiments. 

\subsubsection{Domain-Adversarial Training.}
Every training iteration of the DANN models involves two passes.
The first pass is performed in the same manner as the baseline training procedure and it involves the update (\ref{eq:updateMitosis}).
The second pass uses batches balanced over the domains of the training set, and is used for updates (\ref{eq:updateDomains}) and (\ref{eq:updateAdversarial}).
Given that the training set includes eight domains, the batches for the second pass are therefore made of 8 random patches from each training case.
The learning rate $\lambda_{D}$ for these updates was fixed at 0.0025.

As remarked in \cite{ganin2016domain,kamnitsas2016domain}, domain-adversarial training is an unstable process.
Therefore we use a cyclic scheduling of the parameter $\alpha$ involved in the adversarial update (\ref{eq:updateAdversarial}).
This allows alternating between phases in which both branches learn their respective tasks without interfering, and phases in which domain-adversarial training occurs.
In order to avoid getting stuck in local maxima and to ensure that domain information is not recovered over iterations in the main branch,
the weights of the domain classifier $\boldsymbol{\theta}_{D}$ are reinitialized at the beginning of every cycle.

\subsubsection{Performance.}
The \fscore{1}s for all three approaches and their combinations are given in Table \ref{table:fscores}.
t-SNE embeddings of the feature representations learned by the baseline model and the three investigated approaches are given in Fig. \ref{fig:tsne}.
Although the t-SNE embeddings of the first row only show two domains for clarity, the same observations can be made for almost all pairs of domains.

\begin{table}
\centering
\caption{
Mean and standard deviation of the \fscore{1} over the three repeated experiments.
Every column of the table represents the performance of one method on the internal test set (ITS; from the same pathology lab) and the external test sets (ETS; from different pathology labs).
The squares indicate the different investigated methods.
Multiple squares indicate a combination of methods.
}

\newlength{\cellWidth} \setlength{\cellWidth}{0.10\textwidth}
\newlength{\cellJump} \setlength{\cellJump}{1pt}
\newcommand{\cellTick}{\multicolumn{1}{c}{\rule{6pt}{6pt}}}
\renewcommand{\arraystretch}{1.1}
\newcommand{\cvalue}[2]{\scriptsize $#1 \pm #2$}
\newcommand{\cbvalue}[2]{\scriptsize $\mathbf{#1} \pm \mathbf{#2}$}

\begin{tabular}{c  p{\cellWidth}  p{\cellWidth}  p{\cellWidth}  p{\cellWidth}  p{\cellWidth}  p{\cellWidth}  p{\cellWidth}  p{\cellWidth} }
\hline\noalign{\smallskip}
\textbf{CA}   &   & \cellTick &           &           & \cellTick & \cellTick &           & \cellTick \\
\textbf{SN}   &   &           & \cellTick &           & \cellTick &           & \cellTick & \cellTick \\
\textbf{DANN} &   &           &           & \cellTick &           & \cellTick & \cellTick & \cellTick \\\hline\noalign{\smallskip}

\multicolumn{1}{c}{\ ITS\ } & \cvalue{.61}{.02} & \cvalue{.61}{.01} & \cvalue{.57}{.06} & \cvalue{.61}{.02} & \cvalue{.55}{.01} & \cbvalue{.62}{.02} & \cvalue{.61}{.01} & \cvalue{.57}{.01} \\
\multicolumn{1}{c}{\ ETS\ }  & \cvalue{.33}{.08} & \cvalue{.58}{.03} & \cvalue{.46}{.02} & \cvalue{.55}{.05} & \cvalue{.48}{.08} & \cbvalue{.62}{.00} & \cvalue{.51}{.02} & \cvalue{.53}{.03} \\\hline\noalign{\smallskip}
\end{tabular}
\label{table:fscores} 
\end{table}

\newlength{\subfigWidth} \setlength{\subfigWidth}{0.23\textwidth}
\begin{figure}[!h]
\label{fig:tsne}\centering
\begin{minipage}[c]{\subfigWidth}
	\centering
	\includegraphics[width=1.0\textwidth]{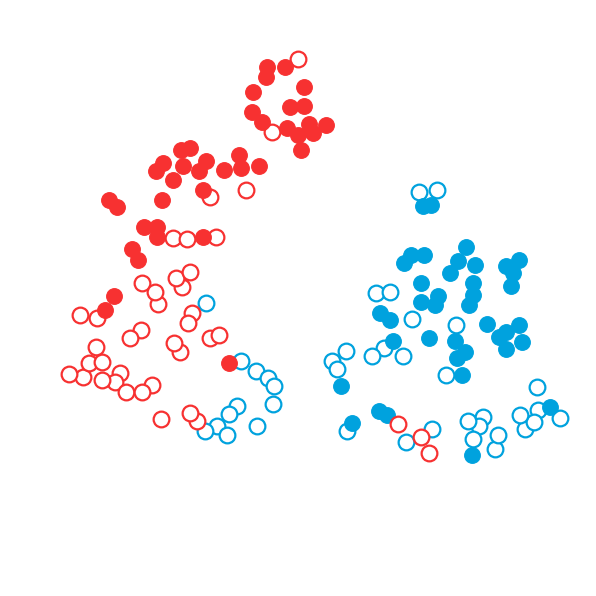}
\end{minipage}
\begin{minipage}[c]{\subfigWidth}
	\centering
	\includegraphics[width=1.0\textwidth]{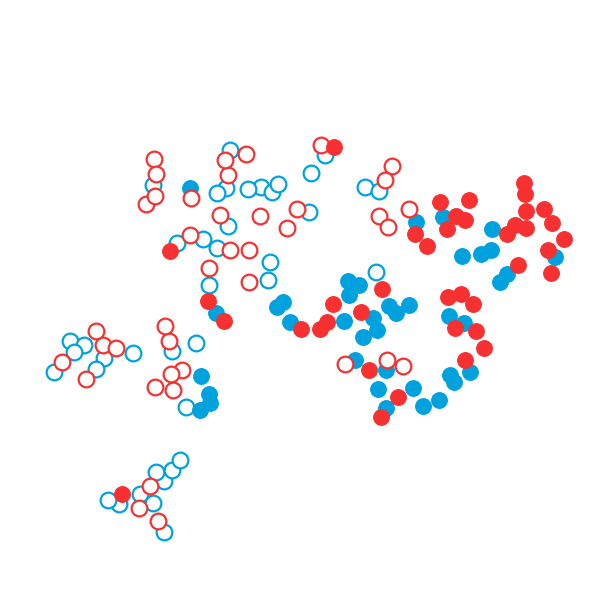}
\end{minipage}
\begin{minipage}[c]{\subfigWidth}
	\centering
	\includegraphics[width=1.0\textwidth]{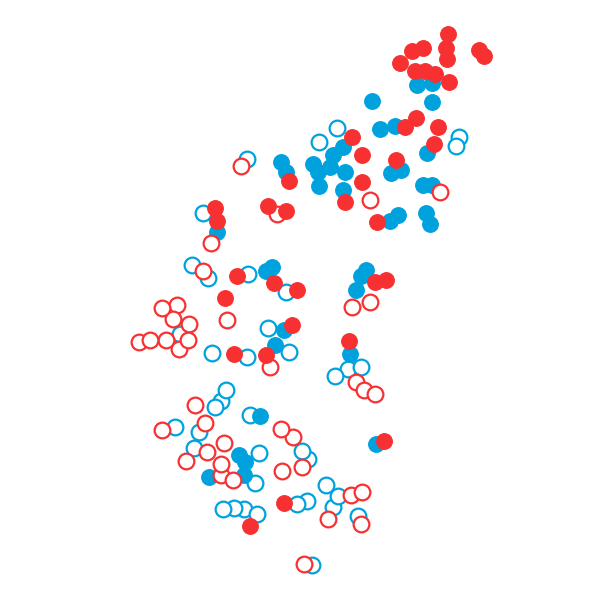}
\end{minipage}
\begin{minipage}[c]{\subfigWidth}
	\centering
	\includegraphics[width=1.0\textwidth]{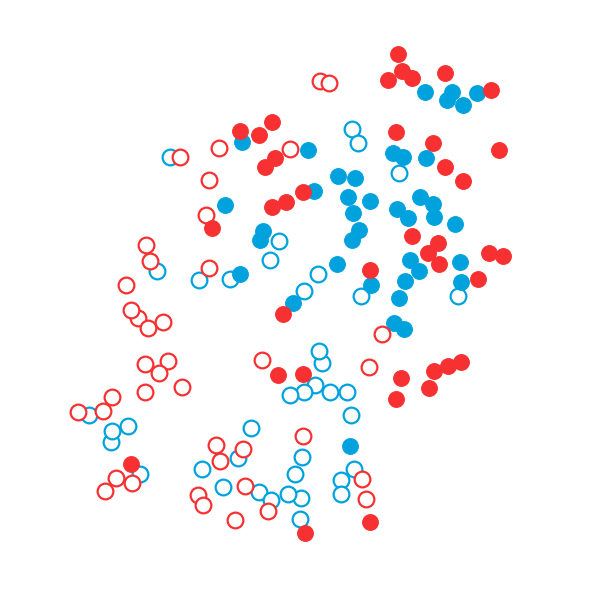}
\end{minipage}
\begin{minipage}[c]{\subfigWidth}
	\centering
	\includegraphics[width=1.0\textwidth]{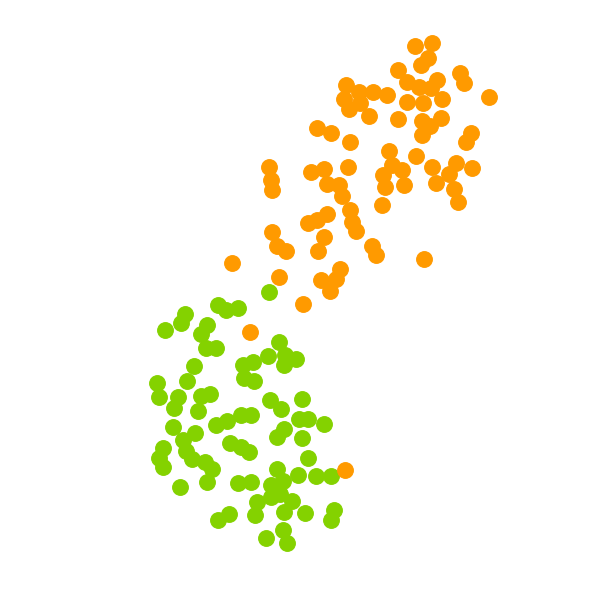}
	(a)
\end{minipage}
\begin{minipage}[c]{\subfigWidth}
	\centering
	\includegraphics[width=1.0\textwidth]{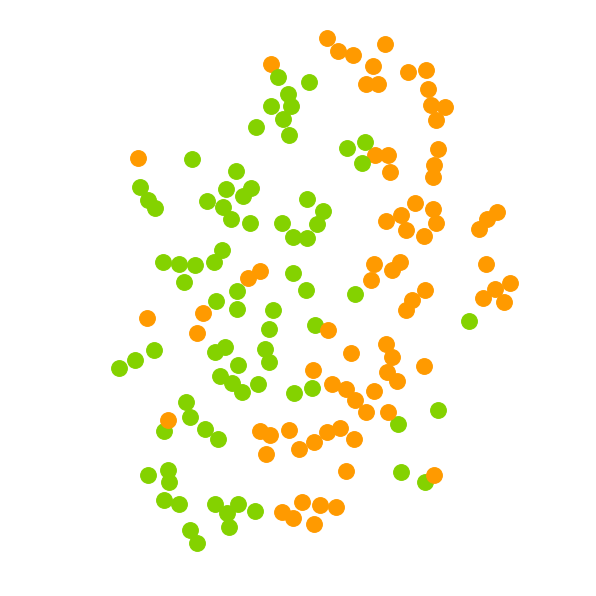}
	(b)
\end{minipage}
\begin{minipage}[c]{\subfigWidth}
	\centering
	\includegraphics[width=1.0\textwidth]{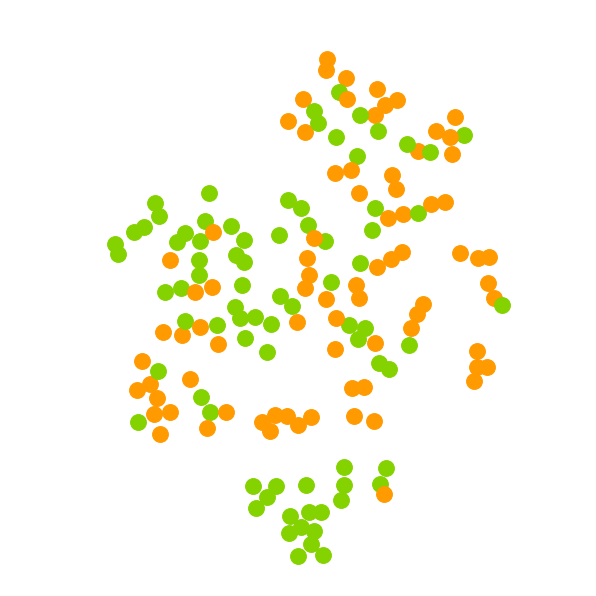}
	(c)
\end{minipage}
\begin{minipage}[c]{\subfigWidth}
	\centering
	\includegraphics[width=1.0\textwidth]{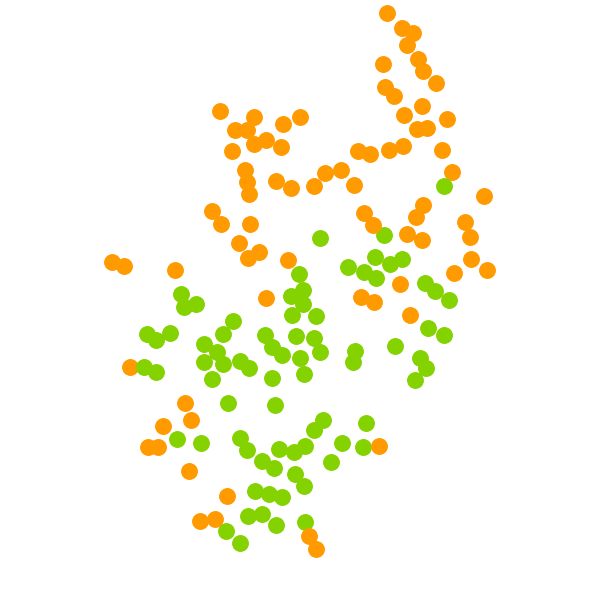}
	(d)
\end{minipage}
\caption{
t-SNE embeddings of 80 patches represented by the learned features at the fourth layer of the mitosis classifier.
First row: patches are balanced across classes (mitosis: disk, non-mitosis: circle) and are equally sampled from two different slides of the training set (red/blue). 
Second row: patches of mitotic figures sampled from slides of the internal (orange) and external test set (green).
Each column corresponds to one approach: (a) baseline, (b) SN, (c) CA, (d) DANN.}
\end{figure}

\section{Discussion and Conclusions}
On the internal test set, all methods and combinations have good performance in line with previously reported results \cite{cirecsan2013mitosis,veta2015assessment,veta2016mitosis,tupac2016}.
The combination of color augmentation and domain-adversarial training has the best performance (\fscore{1} of $0.62 \pm 0.02$).
The staining normalization method and combinations of staining normalization with other methods have the worst performance (\fscore{1}s lower than the baseline method).

As with the internal test set, the best performance on the external test set is achieved by the combination of color augmentation and domain-adversarial training (\fscore{1} of $0.62 \pm 0.00$).
On the external test set, all three investigated methods show improvement since the baseline method has the worst performance (\fscore{1} of $0.33 \pm 0.08$).

The intra-lab t-SNE embeddings presented in the first row of Fig. \ref{fig:tsne} show that the baseline model
learns a feature representation informative of the domain, as shown by the presence of well-defined clusters corresponding to the domains of the embedded image patches.
In contrast, each of the three approaches produces some domain confusion in the model representation, since such domain clusters are not produced by t-SNE under the same conditions.

While staining normalization improves the generalization of the models to data from an external pathology lab,
it clearly has a general adverse effect when combined to other methods, compared to combinations without it.
A possible reason for this effect could be that by performing staining normalization,
the variability of the training dataset is reduced to a point that makes overfitting more likely.

For both test datasets, the best individual method is color augmentation.
The t-SNE embeddings in the second row of Fig. \ref{fig:tsne} show that the models trained with CA
produce a feature representation more independent of the lab than the baseline, SN or DANN.
This is in line with the observation that the appearance variability in histopathology images is mostly manifested as staining variability.

The best performance for both datasets is achieved by the combination of color augmentation and domain-adversarial training.
This complementary effect indicates the ability of domain-adversarial training to account for sources of variability other than color.

In conclusion, we investigated DANNs as an alternative to standard augmentation and normalization approaches, and made a comparative analysis.
The combination of color augmentation and DANNs had the best performance, confirming the relevance of domain-adversarial approaches in histopathology image analysis.
This study is based on the performances for a single histopathology image analysis problem and only one staining normalization method was investigated.
These are limiting factors, and further confirmation of the conclusions we make is warranted.

%
%
\bibliographystyle{splncs03}
\bibliography{references}

\end{document}